\relax
\documentclass[10pt,twocolumn]{article} 
\usepackage{arxiv-style}
\usepackage{times}  
\usepackage{helvet}  
\usepackage{courier}  
\usepackage[hyphens]{url}  
\usepackage{graphicx} 
\usepackage{caption} 
\DeclareCaptionStyle{ruled}{labelfont=normalfont,labelsep=colon,strut=off} 
\usepackage{algorithm}
\usepackage{algorithmic}

\usepackage{amsmath,amssymb}
\usepackage{xcolor}
\usepackage{soul}
\usepackage{adjustbox}
\usepackage{multirow}
\usepackage{amssymb}
\usepackage{amsthm}

\renewcommand{\qedsymbol}{\rule{1.5ex}{1.5ex}}

\usepackage{newfloat}
\usepackage{listings}
\floatstyle{ruled}
\newfloat{listing}{tb}{lst}{}
\floatname{listing}{Listing}
\title{\textit{The King is Naked}: on the Notion of Robustness for Natural Language Processing
}
\author {
    Emanuele La Malfa,
    Marta Kwiatkowska \\
     Department of Computer Science\\ University of Oxford\\
     \{emanuele.lamalfa, marta.kwiatkowska\}@cs.ox.ac.uk
}

\usepackage{bibentry}

\begin{document}

\maketitle

\begin{abstract}
There is growing evidence that the classical notion of adversarial robustness originally introduced for images has been adopted as a \textit{de facto} standard by a large part of the NLP research community.
We show that this notion is problematic in the context of NLP as it considers a narrow spectrum of linguistic phenomena. In this paper, we argue for \textit{semantic robustness}, which is better aligned with the human concept of linguistic fidelity. We characterize \textit{semantic robustness} in terms of biases that it is expected to induce in a model. We study semantic robustness of a range of \textit{vanilla} and robustly trained architectures using a template-based generative test bed.
We complement the analysis with empirical evidence that, despite being harder to implement, \textit{semantic robustness} can improve performance 
on complex linguistic phenomena where models robust in the classical sense fail.
\end{abstract}

\section{Introduction}
In the last decade, deep learning has become the gold standard 
method to solve complex problems in Natural Language Processing (NLP)~\cite{brown2020language}. The  range of NLP applications encompasses text classification~\cite{liu2019roberta}, language translation~\cite{liu2020very}, and now also ranking systems and large-scale search engines~\cite{wang2021dcn}. With models whose complexity -- and consequently, size -- has become `gargantuan'\footnote{GPT-3's full version has 175 billion learning parameters.}, there is an increasing concern about  reproducibility~\cite{liu2021makes} and reliability of those models~\cite{song2020universal}, as it is known that, even for smaller networks, it is possible to exploit their brittleness with techniques of 
adversarial machine learning~\cite{zhang2020adversarial}.
Consequently, concepts of robustness have been transferred from adversarial learning to NLP, resulting in techniques and tools~\cite{li2020adversarial}
that typically check that the network's decision is invariant to a simple bounded perturbation (word substitution or deletion) for a given input (local robustness), working in the (continuous) embedding space or the (discrete) word neighbourhood. However, NLP still lacks a definition of robustness that properly captures linguistic phenomena and is aligned with human common sense~\cite{xu2020elephant}. 
There is currently a debate in the NLP community about the internal working of language models, with some believing they are the `foundation' for the entire discipline~\cite{bommasani2021opportunities} and others arguing that they mostly learn higher-order distributions of words frequency~\cite{sinha2021masked}.
\newline
\indent In this work, we first review the classical notions of robustness adopted in NLP and identify their weaknesses, in terms of the lack of expressiveness and over-reliance on the neural model text representation. Next,  to better align the perception of human robustness to that implemented by a neural model, we formalise (local) \textit{semantic robustness} of NLP as a notion  
that generalizes local discrete robustness through measuring robustness to linguistic rules, rather than to word substitution or deletion. This allows us to define (global) semantic robustness for a linguistic task such as sentiment analysis, which can be extended to higher-order tasks. 
We contribute to the debate in the NLP community by performing a systematic comparison, 
complemented by 
an evaluation of different architectures, of the classical notions of robustness in NLP. We further show that with \textit{semantic robustness} we can evaluate the performance of a model on 
cogent linguistic phenomena, which are of interest for both the NLP and the linguistics community. We achieve this by proposing an assessment framework and a simple, yet effective, test bed based on data augmentation.
Last but not least, we 
wish to highlight the issue of NLP robustness,
which for the last few years has over-focused 
on trivial and often machine-centric symbol manipulation. Using the terminology drawn from cyber-security, this work is a `purple-team' effort to align the key performance indicators of the `red-team' -- whose role is to exploit NLP models with any kind of vulnerability -- with those of the `blue-team', a.k.a. the defenders, who aim to adopt 
a semantic notion of robustness that implies robustness to linguistic phenomena.

\section{Related Work}
Brittleness of neural network models is a serious concern, both theoretically~\cite{biggio2013evasion,szegedy2014intriguing} and practically, including Natural Language Processing (NLP)~\cite{belinkov2018synthetic,ettinger2017towards,gao2018black,jia2017adversarial,liang2017deep,zhang2020adversarial} and more recently complex Masked Language Models (MLM)~\cite{li2020bert,sun2020adv}. In NLP, attacks are usually conducted either at character or word level~\cite{ebrahimi2017hotflip,cheng2018seq2sick}, or at the embedding level, exploiting (partially or fully) vulnerabilities in the symbols' representation~\cite{alzantotgenerating:2018,la2021guaranteed}. Brittleness of NLP does not pertain only to text manipulation, but also includes attacks and complementary robustness for ranking systems~\cite{goren2018ranking}.
Neural network robustness naturally complements the perspective offered by brittleness as it involves the certification of a model against a wide range of attacks~\cite{huang2017safety}. In NLP, similarly to computer vision~\cite{akhtar2018threat}, the majority of works
have adopted the narrow 
notion of robustness, in terms of invariance to minor perturbations of an input text~\cite{gowal2018effectiveness,jia2019certified,dong2021towards,la2020assessing}, 
while only a minority 
have contested these limitations, either implicitly~\cite{ribeiro2020beyond} or explicitly~\cite{morris2020second,morris2020reevaluating,xu2020elephant}, mainly due to the 
difficulty 
of automatically generating semantically involved test beds~\cite{feng2021survey}. Although adversarial data augmentation in NLP is well established~\cite{morris2020textattack}, robustness to semantically coherent, yet possibly diverging, examples is still in its `adolescence'~\cite{ribeiro2018semantically}, as many highly accurate NLP models cannot recognize cogent linguistic phenomena even on low-order tasks such as binary classification~\cite{barnes-etal-2019-sentiment}.

\section{NLP Robustness: a Tale of Two Perspectives}
In this section we discuss the merits of local robustness in NLP, analysing existing concepts and highlighting the issues with local continuous robustness. 
We then introduce the notion of semantic robustness, aimed to better align the perception of human robustness to those implemented by neural models, together with an assessment framework. In the next section, we complement the methodology part with an experimental evaluation, where neural networks' robustness is tested against linguistic phenomena. We complete the paper with a study of the inductive biases that different notions of robustness are expected to induce in a trained model. 

\paragraph{Notation.} We will refer to $f(\cdot)$ as a generic 
neural network that solves a task $T$ for an instance $s$, which is represented as a piece of text written in natural language (e.g., sentiment analysis). W.l.o.g., we will assume that texts are represented as lists of words (features), 
namely $s=(v_1, .., v_l)$. We will denote with $x \in \mathbb{R}^{ld}$ a text $s$ whose l features have been mapped to vectors of real numbers through an embedding, $\mathcal{E}: \mathcal{V} \rightarrow \mathbb{R}^d$, where $\mathcal{V}$ is a finite vocabulary of words. We refer to a component of $x$ along a generic embedding axis as $x^{(i)} \in \mathbb{R}, \ i \in \ \{1, ..,d\}$. Since the majority of the embedding spaces are injective non-surjective functions, the notation $x_v=(x_{v_1}, .., x_{v_l})$ will serve to denote an embedded text $x$ that further admits, for each vector $x_{v_i}$, a preimage in the vocabulary space, i.e., $\forall x_{v_i} \in x_v \ \exists! \ v \in \mathcal{V} \ . \ \mathcal{E}(v)=x_{v_i}$. With a slight abuse of notation, we will denote with $\mathcal{E}(s)=(x_{v_1}, .., x_{v_l})$ a text whose words have each been embedded through $\mathcal{E}$. Finally, we will assume that the first operation of the model $f(\cdot)$ involves a transformation  
through an embedding representation $\mathcal{E}$. 

\subsection{Classical Notions of NLP Robustness}
We begin by discussing 
the concept of local continuous robustness, which is widely used in computer vision 
and has been applied to NLP~\cite{huang2019achieving,jia2017adversarial,la2020assessing}. We then consider local discrete robustness, which manipulates symbols rather than embedding vectors~\cite{alzantotgenerating:2018}. We show that the former notion can be reduced to the latter: nonetheless, both definitions only  allow one to express robustness to a limited number of linguistic phenomena. We extensively discuss the advantages and drawbacks 
of those two notions.

\paragraph{Definition 1 (Local Continuous Robustness).} A model $f(\cdot)$ is locally robust to $\epsilon$-bounded perturbations when, given a task $T$ and one of its instances $x$, it holds that $\forall x' \in Ball_{\epsilon}(x), \ f(x)=f(x')$, where $Ball_{\epsilon}(x)=\{x' \ . \ ||x-x'||_p\le\epsilon\}$, $||_p$ is an $L_p$ norm of choice and $\epsilon\ge0$ a (small) real number. 

\begin{figure}[t]
    \centering
    \includegraphics[height=0.6\linewidth,width=0.6\linewidth]{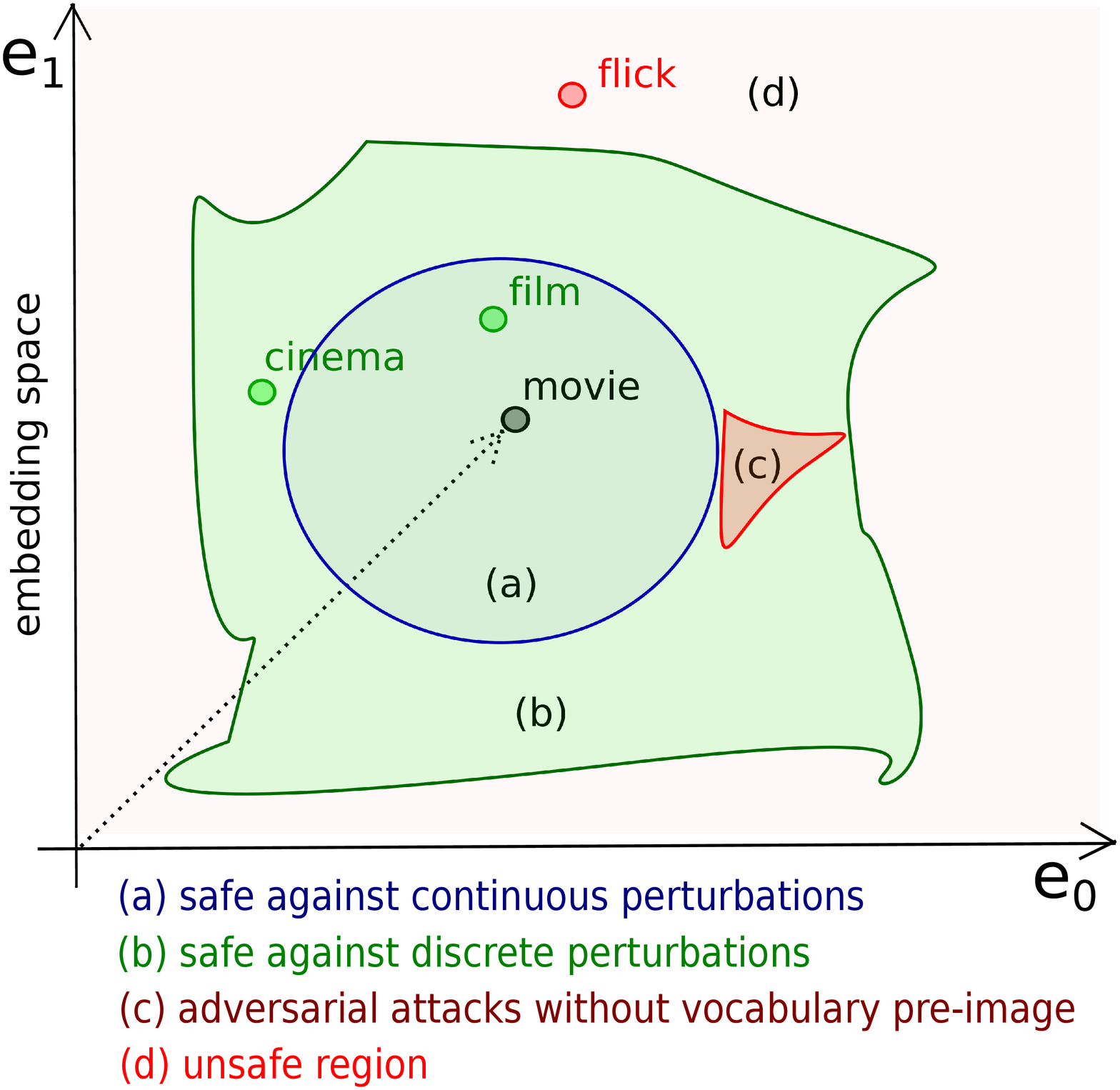}
    \caption{
    In general, local continuous robustness is an ill-posed property for NLP. A model can be robust to a large surface of attacks in the input neighbourhood (green patch (b)), yet a small region of adversarial attacks (red patch (c)) invalidates the verification of larger regions. In the example, the safe input neighbourhood (blue patch (a)), a convex region that includes safe replacements, cannot grow any further without violating robustness by encroaching on patch (c). 
    Non-convex representations for an input neighborhood (patch (a)) are possible, but computationally expensive and not 
    used in practice.}
    \label{fig:discrete-vs-continuous}
\end{figure}

\paragraph{Observation 1.} \ul{Natural language is discrete while local continuous robustness is defined over a dense representation.} Standard embedding techniques~\cite{mikolov2013distributed,gloveemb:2020} define the word-to-vector mapping over a continuous space, with the input vocabulary discrete and finite (e.g., characters, words, sentences) and the output dense and uncountable. On the other hand, natural language is discrete and allows for finite, yet combinatorial, outcomes. In this \textit{hybrid} setting,  $\epsilon$-bounded robustness implies that any vector in this dense $\epsilon$-bounded region is safe. This assumption is linguistically inconsistent, as a network may present a decision boundary where an adversarial attack that is not a proper word limits the verification or severely reduces 
the safe region. We illustrate this issue in Figure~\ref{fig:discrete-vs-continuous}. 

\begin{figure}[t]
    \centering
    \includegraphics[width=0.7\linewidth]{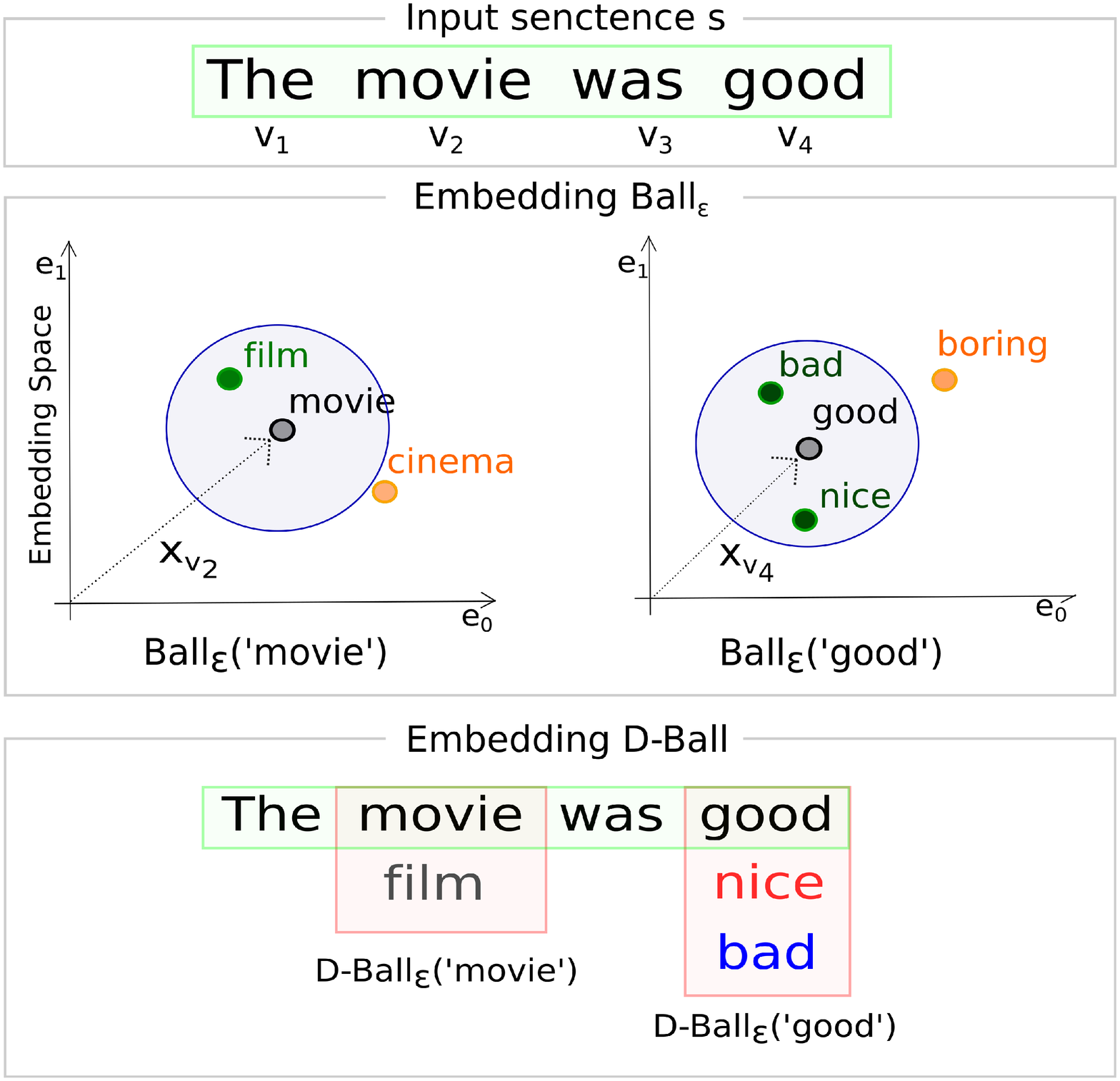}
    \caption{
    $Ball_{\epsilon}$ (top) and $D$-$Ball_{\epsilon}$ (bottom) representations of two words from the input sentence \textit{``the movie was good''} ($s$ in our notation). For the same value of $\epsilon$, $Ball_{\epsilon}$ contains all the discrete replacements of the equivalent $D$-$Ball_{\epsilon}$ plus all the vectors (infinitely many) that cannot be mapped back to the vocabulary  $\mathcal{V}$ (inside each blue ball around an input word).}
    \label{fig:ball-dball}
\end{figure}


\paragraph{Definition 2 (Local Discrete Robustness).} A model $f(\cdot)$ is locally robust to discrete perturbations when, given a task $T$ and an instance $x_v$ embedded from $s$, it holds that $\forall x_v' \in D$-$Ball_{\epsilon}(x_v), \ f(x_v)=f(x_v')$, where $D$-$Ball_{\epsilon}(x_v)=\mathcal{E}(\mathcal{V})^l \cap Ball_{\epsilon}(x_v)$.
\newline \newline
We exemplify the differences between $Ball_{\epsilon}$ and the corresponding $D$-$Ball_{\epsilon}$ in Figure \ref{fig:ball-dball}.

\paragraph{Proposition 1.}
\ul{Local continuous robustness implies local discrete robustness, but the converse is generally false.} \newline
\textbf{Proof.} From a mathematical perspective, 
$D$-$Ball_{\epsilon}(x) \subseteq Ball_{\epsilon}(x)$ but the opposite is not true. 
For $\epsilon=0$, both $D$-$Ball_{\epsilon}$ and $Ball_{\epsilon}$ are singletons. \qedsymbol 

\paragraph{Observation 2.}\ul{Continuous and discrete robustness allow but 
limited syntax manipulations.} As demonstrated empirically by many existing works in the literature \cite{alzantotgenerating:2018,jia2019certified,huang2019achieving,dong2021towards}, both formulations of robustness only allow for robustness testing  against symbol-to-symbol substitutions or deletions. The limited degree of freedom of an operator that  locally substitutes a word with other words makes it hard, if not impossible, to test for robustness against 
paraphrases. As an example, if a model $f(\cdot)$ is robust for the sentence \textit{``the movie was good''}, 
which implies  correct classification for the texts \textit{``the film was good''}, \textit{``the film was nice''}, etc., we cannot say the same for the sentence \textit{``an enjoyable thriller''}. From the linguistic perspective, this problem arises since the frequency of words in natural language follows the Zipf's law~\cite{zipf2013psycho}, where rare terms and constructs -- hence edge cases -- occur more frequently than in other natural phenomena.

\paragraph{Observation 3.} \ul{There is no guarantee that perturbations in both discrete and continuous settings do not \textit{violate} the task under consideration.} As the methods that implement both discrete and continuous robustness allow for weak supervision in the choice of the replacements, a perturbation 
can diverge from the task under consideration. It is well known that many recent embeddings have been developed to be faithful to a (static) version of the ``distributional hypothesis''~\cite{baroni2014don}, and thus it is not unusual to find words like \textit{``bad''} and \textit{``good''} close to each other in the representations. This could lead to potentially disastrous 
effects when balancing local robustness, e.g.,~\cite{gowal2018effectiveness}, with accuracy, especially for low-order tasks such as sentiment analysis.

\subsection{A Semantic Notion of Robustness}
We now introduce a notion of robustness that goes beyond word replacements, and thus permits an assessment of the brittleness to linguistic phenomena that are cogent to humans. To do so, we first need to introduce some notation.

\paragraph{Definition 3 (Oracle).} An Oracle $\Omega$ solves a task $T$ for any input $s$ that is compliant with $T$, while it rejects all those instances that are not. We denote with $\Omega \models s$ the act of solving a task, and with $\Omega \not \models s$ the rejection. Solving and rejection are mutually exclusive. 

\paragraph{Observation 4.} \ul{An Oracle is an \textit{augmented, idealized} linguistic model.} There are two cogent differences between an Oracle $\Omega$ and a standard model $f(\cdot)$: (i) a model can be wrong on samples from $T$ (e.g., misclassifications) while the Oracle is always right about its decisions; (ii) the Oracle (certainly) rejects inputs that are not compliant with the task $T$.

\paragraph{Example 1.} \ul{An Oracle for sentiment analysis.} Given a sentiment analysis task $T$ for movie reviews, an Oracle $\Omega$ correctly classifies any text that expresses a judgment about the movie. As an example, \textit{``the movie was (not) good''} will be classified as positive (negative). An Oracle alternatively rejects any piece of text that is inconsistent with $T$, i.e., all those texts that do not explicitly (or implicitly) express a judgment about a movie. An example is the text \textit{``recipe of risotto with mushrooms: [...]''}, which is rejected. From a practical perspective, a classifier $f(\cdot)$ admits misclassifications in the sense that its accuracy may not be maximal (i.e., less than $1.$) and it further cannot reject inputs that are not compliant with the task: in the case it does, the task is not fulfilled perfectly.  

\paragraph{Definition 4 (Linguistic Rule).} A linguistic rule is a symbolic function that 
manipulates a text $s$ according to a linguistic phenomenon and a task $T$, whose generated texts $S'$, along with the original input, are not rejected by $\Omega$. Formally, $R: (s,T) \mapsto S' \ . \ \forall \ s' \in S', \ \Omega \models  s \wedge s'$. 

\paragraph{Observation 5.} \ul{Linguistic rules are flexible symbolic methods.} 
Since a linguistic variation of an input can be very different from the original text, a rule should be allowed to add/remove/replace words while remaining compliant with the task $T$. As an example of a simple rule, one can think of \textit{verb negation} that acts on a text and negates the action expressed by the subject (if any). While this task is often trivial for humans, fully algorithmic solutions to this problem are still limited in their capabilities \cite{guo2018long}. \textit{Hybrid} methods, based on synthetic data augmentation, humans-in-the-loop and deep MLM \cite{feng2021survey,lin2019commongen,huang2019reducing}, constitute currently an active area of NLP research. One viable way to generate the replacements is to use template-based data-augmentation techniques (as employed in this paper and detailed in Experimental Evaluation). More complex approaches involve MLM with humans-in-the-loop who validate the generated perturbations. While for the generative process the MLM can be trained to be controlled through textual `seeds' (e.g., in the spirit of the works by \cite{wu2021polyjuice} or \cite{madaan2020generate}), humans play the role of the Oracle.

\paragraph{Example 2.} \ul{A rule for \textit{shallow negation}.}
For a sentiment analysis task $T$ with positive and negative instances and a positive instance $s$ \textit{``the movie was good''}, the \textit{shallow negation} rule R negates the sentiment expressed by $s$, and hence valid perturbations generated by $R$ on $s$ are \textit{``the movie was not good''}, \textit{``a bad film''}, but also more involved examples like \textit{``it is false that the movie is good''}, etc. We name this rule shallow negation as it does not allow for nested negations, regardless of their grammatical consistency (i.e., \textit{``it is false that the movie wasn't good''} cannot be generated by $R$ on $s$).

\paragraph{Definition 5 (Local Semantic Robustness).} Formally, given a model $f(\cdot)$, a linguistic rule R, an input $s$ from a task $T$, a measure of the performance of $f(\cdot)$ on $T$, namely $p \in [0, 1]$ (with 0. denoting a random guess and 1. perfect accuracy), a small number $\tau\ge0$, and a measure of performance $p'$ on samples $S'$ generated by $R(s,T)$, we say $f(\cdot)$ is $\tau$-semantically robust for $R$ and $s$ if it holds that $\mathbb{E}_{s' \sim S'}[p'] \ge p - \tau$, with 
$min(0,p-\tau)=0$. 
\newline


We further say that a model is \textit{bounded invariant} to a rule $R$ when it holds that $p-\tau \le \mathbb{E}_{s' \sim S'}[p'] \le p+\tau$. 

\indent Informally, a model $f(\cdot)$ that correctly classifies an instance $s$ of a task $T$ is semantically robust to a linguistic rule $R$ when it exhibits at least the same performance on the set $S'$ of perturbations $s'$ generated by applying $R$ to $s$.  We further observe that this formulation allows for the performance $p'$ to even surpass $p$, so this notion entails that $f(\cdot)$ is no worse at correctly solving $T$ for $S'$ than it is at solving any other task, and is hence a stronger notion 
than \textit{bounded invariance}.

\paragraph{Observation 6.} \ul{Local semantic robustness is linguistically meaningful.} The notion is local as linguistic rules act on a single text $s$. It is further inherently linguistic as the transformation $R$ of an input text acts at the syntax level but then the Oracle's reject phase guarantees it has preserved the semantics of each $s'$ w.r.t. $T$.

\paragraph{Observation 7.} \ul{Local semantic robustness is entailed by linguistic generalization, but not the other way round.} Linguistic robustness is different from generalization on unseen test cases. The former is entailed by the latter, while the other way round is not necessarily true. Semantic robustness is defined over a rule while generalization is a more general and hard to obtain/optimize objective.

\paragraph{Proposition 2.} \ul{Local semantic robustness can be reduced to local discrete robustness, but not to local continuous robustness.}
\newline
\textbf{Proof.} For local discrete robustness, it is straightforward to define a rule that generates perturbations according to the definition of local discrete robustness. In this sense the semantic rule R involves extracting the replacements in the embedding's neighborhood of each input word. \newline 
As regards local continuous robustness, the invariance over all the input texts $s'$ in an $\epsilon$-ball cannot be mapped back to the embedding (a.k.a. input) vocabulary $\mathcal{V}$ by any combination of linguistic rules as they act, by definition, at the symbol level. Since the majority of continuous embeddings are injective non-surjective functions, almost all the vectors in any non-empty region of the space cannot be mapped back to a proper entry of $\mathcal{V}$. \qedsymbol  

\paragraph{Definition 6 (Semantic Robustness).}
This notion extends local semantic robustness beyond the single instance and to a specific task T. A model $f(\cdot)$ exhibits global semantic robustness (or in general semantic robustness) to a rule R and a task $T$ when it is locally semantically robust for any input $s'$ generated by applying R to a test set. 

\paragraph{Assessment Framework for Semantic Robustness} 
A sufficient condition for quantifying the semantic robustness of a model on an NLP task is that it is possible to measure the performance of such a model on unseen input texts.
In this sense, we can measure the semantic robustness of a model $f(\cdot)$ that solves a task T by comparing its performance $p$ with the performance $p'$ of the model on an unseen test bed that contains one or more semantic phenomena.

We now describe some illustrative examples of measuring semantic robustness, firstly for sentiment analysis and then for more involved NLP tasks.

\paragraph{Example 3.}\ul{Robustness to \textit{shallow negation} in sentiment analysis.}
Given a sentiment analysis task with positive and negative instances, a model $f(\cdot)$ trained on a dataset $(S,Y)$ and validated on $(S_{test}, Y_{test})$ is robust to shallow negation when $\forall s \in S_{test}, \ \forall s' \in S'=R(s,T), \ (\Omega \models s \wedge s') \Rightarrow \mathbb{E}_{s' \sim S'}[p'] \ge p - \tau$ for some $\tau\ge0$, with $R$ the negation rule that acts on a specific text 
and negates the sentiment expressed by $s$. In this sense, $p$ represents the accuracy of the trained model on $(S_{test}, Y_{test})$, while $p'$ is the accuracy measured on a subset of samples that contain specific linguistic phenomena. We remark that a test bed can be handcrafted, as we show in our paper, or distilled from existing datasets, as described in \cite{barnes-etal-2019-sentiment}.  

\paragraph{Example 4.}\ul{Semantic robustness in high-order NLP tasks.}
We now briefly sketch how we would approach the measurement of semantic robustness for higher-order NLP tasks.
For Question and Answer (QA) tasks, a measure of robustness  can be quantified as the gap between the `unexpectedness` of an Answer when the Question does/doesn't contain a linguistic phenomenon. 
In Natural Language Inference (NLI), directly applying our framework would be straightforward since NLI is reducible to a classification task. In the same way, when Read and Comprehension (RC) is pursued in the form of a classification task, the evaluation of semantic robustness would be similar to sentiment analysis or NLI, whereas when the answer requires re-elaborating the input, the measurement of semantic robustness would be similar to QA (with possibly a different evaluation metric for T).

\begin{table*}[t]
\centering
\begin{tabular}{|l|l|l|l|l|l|}
\hline
\textbf{} & \textbf{Train} & \textbf{FCs} & \textbf{CNNs} & \textbf{LSTMs} & \textbf{Self-attention} \\ \hline
\textit{\textbf{Shallow Negation}} &
  \textit{\begin{tabular}[c]{@{}l@{}}Vanilla\\ Augmented\end{tabular}} &
  \begin{tabular}[c]{@{}l@{}}\small{0.4034} $\pm$ \small{0.0214}\\ \small{0.4062} $\pm$ \small{0.0167}\end{tabular} &
  \begin{tabular}[c]{@{}l@{}}$0.4032 \pm 0.0124$\\ $0.4249 \pm 0.0255$\end{tabular} &
  \begin{tabular}[c]{@{}l@{}}$0.4771 \pm 0.0143$\\ $0.6387 \pm 0.0387$\textbf{*}\end{tabular} &
  \begin{tabular}[c]{@{}l@{}}$0.4790 \pm 0.0059$\\ $0.5954 \pm 0.0027$\textbf{*}\end{tabular} \\ \hline
\textit{\textbf{Mixed Sentiment}} &
  \textit{\begin{tabular}[c]{@{}l@{}}Vanilla\\ Augmented\end{tabular}} &
  \begin{tabular}[c]{@{}l@{}}$0.4707 \pm 0.0360$\\ $0.4912 \pm 0.0339$\end{tabular} &
  \begin{tabular}[c]{@{}l@{}}$0.4986 \pm 0.0415$\\ $0.5271 \pm 0.0387$\end{tabular} &
  \begin{tabular}[c]{@{}l@{}}$0.5110 \pm 0.0251$\\ $0.6357 \pm 0.0317$\textbf{*}\end{tabular} &
  \begin{tabular}[c]{@{}l@{}}$0.5487 \pm 0.0099$\\ $0.5617 \pm 0.0048$\end{tabular} \\ \hline
\textit{\textbf{Sarcasm}} &
  \textit{\begin{tabular}[c]{@{}l@{}}Vanilla\\ Augmented\end{tabular}} &
  \begin{tabular}[c]{@{}l@{}}$0.5136 \pm 0.0504$\\ $0.5297 \pm 0.0657$\end{tabular} &
  \begin{tabular}[c]{@{}l@{}}$0.4681 \pm 0.0327$\\ $0.4678 \pm 0.0317$\end{tabular} &
  \begin{tabular}[c]{@{}l@{}}$0.5578 \pm 0.0128$\textbf{*}\\ $0.4807 \pm 0.0197$\end{tabular} &
  \begin{tabular}[c]{@{}l@{}}$0.5240 \pm 0.0132$\\ $0.6236 \pm 0.0218$\textbf{*}\end{tabular} \\ \hline
\end{tabular}
\caption{
Comparison of accuracy of $20$ \textit{vanilla} and \textit{augmented} models obtained for four different architectures (FCs, CNNs, LSTMs and self-attention), on three linguistic phenomena (\textit{shallow negation}, \textit{mixed sentiment} and \textit{sarcasm}). All the networks have been trained on the SST-2 dataset. \textit{Augmented} models are \textit{vanilla} models fine-tuned on the linguistic rules of interest. Symbol \textbf{*}, when present, means that the improved performance (from \textit{vanilla} to \textit{augmented}, or the other way round) is statistically significant. Interestingly, \textit{sarcasm} is harder to learn and models fine-tuned on this phenomenon perform as well as their \textit{vanilla} counterparts (when not worse).}\label{tab:robustness}
\end{table*}

\section{Experimental Evaluation}
We next conduct an extensive experimental evaluation\footnote{The code for full reproducibility of the experiments is available at \url{https://github.com/EmanueleLM/the-king-is-naked}. All of the experiments have been conducted on a Fedora 32 mid-end laptop equipped with 16GB of RAM and an Intel-i5 CORE $8^{th}$-generation. All the neural network models have been built, trained and tested with Keras~\cite{chollet2015keras}, while for experiments that involved BERT~\cite{devlinbert:2018} we relied on the PyTorch implementation~\cite{NEURIPS20199015}.} designed to answer the following research questions: (i) whether models robust in the classical sense are also semantically robust; (ii) whether robustness to specific linguistic phenomena is a by-product of training accurate NLP classifiers; (iii) whether,  for different architectures, augmented supervised training -- with texts that contain a specific linguistic phenomenon -- induces generalization on unseen test samples that contain the same phenomenon; (iv) whether it is possible to train models that are both accurate and semantically robust, and (v) to what extent unsupervised learning contributes to semantic robustness. 
\newline
We conduct the experiments on models trained -- or fine-tuned through data augmentation -- on the Stanford Sentiment Treebank dataset (SST-2)~\cite{socher2013recursive} and on the dataset collected by~\cite{barnes-etal-2019-sentiment}. The advantages of this approach are two-fold. Firstly, human experts have collected/handcrafted sentences whose syntax/semantics is rich and the level of noise restrained. Secondly, since in NLP spurious patterns and over-fitting play a crucial role during training whose influence is hard to estimate and quantify, cogent compactness of those datasets makes it relatively easy to assess the results. 
To further estimate the robustness on linguistic phenomena, in the spirit of the evaluation done in~\cite{huang2019reducing}, we utilise a template-based method, whose details are given below, for generating augmented samples for a selection of linguistic phenomena to create a test bed, which we use for systematic evaluation of semantic robustness.
In order to validate the soundness of our generative test bed, 
we compare the performance of our rule-generated semantically robust models from our benchmark 
to those examples in~\cite{barnes-etal-2019-sentiment} that exhibit the same linguistic phenomenon, showing comparable accuracy.

\paragraph{Linguistic phenomena.}
Following the work in~\cite{barnes-etal-2019-sentiment}, we have chosen interesting linguistic and para-linguistic phenomena, taking care to exclude those that require external knowledge to be solved (i.e., not explicitly expressed in the sentence). As an example, consider the review \textit{``This movie is another Vietnam''}, which can be correctly classified as negative if the model has some knowledge of that specific way of saying (i.e., exogenous knowledge). 
We now briefly describe the linguistic phenomena that are the object of our robustness evaluation:
\begin{itemize}
    \itemsep0em 
    \item \textit{Shallow negation}: when the sentiment of a sentence is negated. We do not consider nested negations, which make the recognition of the phenomenon considerably harder~\cite{wiegand2010survey,socher2013recursive,prollochs2015enhancing}.
    \item \textit{Mixed sentiment}: when phrases of different polarity appear in the same sentence~\cite{kenyon2018sentiment,barnes-etal-2019-sentiment}. We only consider texts where the overall sentiment is still not ambiguous for a human.
    \item \textit{Irony/sarcasm}: when a sentence makes some premises that are then violated~\cite{hao2010ironic}. This is known to be one of the hardest, yet pervasive, linguistic phenomena of human language.
\end{itemize}

\paragraph{Template-based linguistic rules.}
In addition to the test beds provided by \cite{barnes-etal-2019-sentiment}, in our work we consider a template-based method for generating augmented samples that contain a specific linguistic phenomenon. We pre-define a selection of templates for which we know the corresponding output labels (i.e., \textit{positive} or \textit{negative}). In a template, part of the text is fixed while the remaining part is symbolically represented by tokens which are iteratively replaced by combinations of words from candidate perturbation sets. The augmentation preserves the semantics of the sentence while introducing a linguistic phenomenon (such as \textit{shallow negation}). In our implementation of the rules, a perturbation cannot change the template's label: in this sense, the rejection phase (see Definition 3) is embedded in the generative pipeline, while a process that involves an MLM and generations that are possibly label-changing might be supervised by a human. Examples of templates for each linguistic rule are included in Table~\ref{tab:templates}, along with candidate replacements for each token in Table~\ref{tab:templates-subs}. 

\begin{table}[t]
\centering
\begin{tabular}{ll}
\hline
\multicolumn{1}{c}{\textbf{Tokens}} & \multicolumn{1}{c}{\textbf{Replacements}}          \\ \hline
@NEGATIVE@                          & \textit{'bad', 'poor', 'boring', {[}...{]}}        \\ \hline
@POSITIVE@                          & \textit{'good', 'nice', 'fantastic', {[}...{]}}    \\ \hline
@NAME@                              & \textit{'Uma', 'Bruce', 'Sandra', {[}...{]}}     \\ \hline
@SURNAME@                           & \textit{'Thurman', 'Willis', 'Bullock', {[}...{]}} \\ \hline
@CATEGORY@                          & \textit{'thriller', 'horror', 'comedy', {[}...{]}} \\ \hline
@BOOLFALSE@                         & \textit{'false', 'wrong', 'incorrect', {[}...{]}}  \\ \hline
@AUGMENT@                         & \textit{'very', 'extremely', 'incredibly', {[}...{]}}  \\ \hline
\end{tabular}
\caption{Candidate perturbation sets used to generate combinations of replacements in template-based texts (Table \ref{tab:templates}).}
\label{tab:templates-subs}
\end{table}

\begin{table*}[t]
\centering
\begin{tabular}{ll}
\hline
\multicolumn{1}{c}{\textit{\textbf{Shallow Negation}}}                                  & \textbf{Label}    \\ \hline
'This @CATEGORY@ movie is not @AUGMENT@  @NEGATIVE@.'                                   & \textit{positive} \\ \hline
'It is @BOOLFALSE@ that this @CATEGORY@ movie is @AUGMENT@ @POSITIVE@.'                 & \textit{negative} \\ \hline
\multicolumn{1}{c}{\textit{\textbf{Mixed Sentiment}}}                                   & \textit{}         \\ \hline
'Despite @NAME@ @SURNAME@ acted well,  this @category@ movie is @augment@ @negative@.'  & \textit{negative} \\ \hline
'A @AUGMENT@ @NEGATIVE@ plot for a @AUGMENT@ @POSITIVE@ movie.'                         & \textit{positive} \\ \hline
\multicolumn{1}{c}{\textit{\textbf{Sarcasm}}}                                           & \textit{}         \\ \hline
'Starring @NAME@ @SURNAME@ i would prefer to be killed rather than watching this @CATEGORY@ movie.' & \textit{negative} \\ \hline
'Please throw this @AUGMENT@ long @CATEGORY@ movie into the ocean, and thank me later.' & \textit{negative} \\ \hline
\end{tabular}
\caption{Examples of template-based reviews, along with the ground truth label, used to generate sentences that contain the linguistic phenomena studied in the paper.}
\label{tab:templates}
\end{table*}

\subsection{Comparative Study}
We compare architecturally different models on the 
three linguistic phenomena we previously introduced.
We conduct an extensive evaluation on four 
neural architectures, namely fully connected (FC), convolutional (CNN)~\cite{zhangcharacter:2015}, Long Short-Term Memory (LSTM)~\cite{hochreiter1997long} and self-attention~\cite{vaswani2017attention}. We choose the number of hidden units of each layer so that the number of parameters is approximately the same and in the order of $40$K\footnote{Each input text is $25$ words long (eventually padded or cut), while each word is mapped to a vector of real numbers through a $50$-dimensional embedding, pre-trained on the SST-2 task~\cite{chollet2015keras}. Each network is composed of $3$ layers, where the topology of the last two is shared, i.e., respectively a $32$ hidden units ReLU and a $2$ hidden units softmax layer (both are dense). The first layer depends on the specific topology under examination (e.g., self-attention will have a self-attention layer, LSTM a Long Short-term Memory cell, etc.): the first layer has $32$ hidden units for the FCs, $44$ ReLU kernels of size $3$ for the CNNs, $75$ tanh hidden units for the LSTMs and $32$ ReLU hidden units for the self-attention networks.}. For each linguistic phenomenon, we analyse and compare the robustness of $20$ models trained on \textit{plain} SST-2 dataset (i.e., no semantic data augmentation of any kind) and then on a \textit{semantically augmented} version of the same dataset (details of the augmentation are provided in the relevant subsection).

\paragraph{\textit{Vanilla} Models.}
For each linguistic phenomenon introduced in the previous section, we analyse and compare the robustness of $20$ models trained on 
the SST-2 dataset without augmentation. 
For FCs and CNNs the average accuracy on the SST-2 test set is $0.8993 \pm 0.0029$ and $0.9077 \pm 0.0038$ respectively, while the accuracies of LSTMs and self-attention are $0.9101 \pm 0.0033$ and $0.8963 \pm 0.0015$. We report in Table~\ref{tab:robustness} the results of each population for the three 
linguistic phenomena in this study. Self-attention and RNN-LSTMs are the best performers, while FCs and CNNs have lower accuracy in all the three tasks. Interestingly, none of the models, despite having a high accuracy on the test set, is able to  recognize any linguistic construct we tested. On the one hand, this analysis, which can guide the design when seeking to enforce appropriate inductive biases of a neural architecture \cite{kharitonov2020they}, provides additional evidence for the vast literature on the limitations of accuracy when judging the linguistic performance of an NLP model~\cite{socher2013recursive,kenyon2018sentiment,barnes2021improving}. On the other hand, it motivates our next step, which involves fine-tuning the same architectures on texts that exhibit these linguistic phenomena.

\paragraph{Semantic robustness through data augmentation.}
In this section, we study how -- and to what extent -- data augmentation, along with architectural inductive biases, can be used to inject semantic robustness to different linguistic phenomena. We re-trained the models of the previous section by adding samples from~\cite{barnes-etal-2019-sentiment} that contain one of the specific rules used previously to the training set, up to a multiplicative factor to balance the large number of samples of the SST-2 dataset\footnote{With the SST-2 train set that accounts for approximately $112$K input samples and each semantic rule that generates roughly $500-1000$ new samples, semantic data augmentation with a multiplicative factor of $1$ accounts for additionally $1$K samples, etc.}. 
While for a multiplicative factor of $500$ none of the models exhibit any improvement in the semantic tasks, for a multiplicative factor of $750$ we observe some improvement in LSTMs and self-attention. While the experiments suggest that FCs and CNNs cannot learn any of the three linguistic phenomena we studied, LSTMs and self-attention networks benefit from data augmentation. With reference to Table~\ref{tab:robustness}, both LSTMs and self-attention improve considerably on \textit{shallow negation}. On \textit{mixed sentiment}, augmented LSTMs substantially improve over the \textit{vanilla} counterpart, while self-attention does not seem to exploit the additional information (despite a slight improvement over the \textit{vanilla} case). Finally, data augmentation allows self-attention to improve significantly on \textit{sarcasm}, though the same regime is detrimental for LSTMs, where the \textit{vanilla} networks consistently outperform those trained on augmented data. 
Finally, for a multiplicative factor of $1$K or superior, we observe a detrimental effect on the robustness of each model that is comparable to the \textit{vanilla} SST-2 training. 

\begin{table}[t]
\centering
\begin{adjustbox}{width=1\columnwidth,center}
\begin{tabular}{|l|l|l|}
\hline
\textbf{} & \textbf{\begin{tabular}[c]{@{}l@{}}Accuracy \\ (Barnes et al., 2019)\end{tabular}} & \textbf{\begin{tabular}[c]{@{}l@{}}Accuracy \\ (Our Benchmark)\end{tabular}} \\ \hline
\textit{\textbf{Shallow Negation}} & 0.8552 & 0.7928 \\ \hline
\textit{\textbf{Mixed Sentiment}}  & 0.6024 & 0.6974 \\ \hline
\textit{\textbf{Sarcasm}}          & 0.7111 & 0.8455 \\ \hline
\end{tabular}
\end{adjustbox}
\caption{Summary of BERT semantic robustness on different linguistic phenomena, tested on samples from~\cite{barnes-etal-2019-sentiment} (left column) and from our template-based benchmark (right column). For these results, a BERT model has been fine-tuned on the SST-2 dataset.}
\label{tab:bert}
\end{table}

\subsection{Classic Robustness is Linguistically Brittle}%
We have compared \textit{robust} models trained with IBP (Interval Bound Propagation)~\cite{gowal2018effectiveness} with their \textit{vanilla} counterparts. For different values of $\epsilon=(0.001,0.01)$ in the $L_{\infty}$-norm, which makes the model-to-model results easy to compare \cite{la2020assessing},  
and an embedding diameter of approximately 3.17, we assess IBP-induced robustness on semantic rules. Interestingly, their performance is comparable (when not worse) to the brittle counterparts for all the linguistic phenomena we analyse, thus validating our previous Observation 2, i.e., that models robust in the classical sense have an extremely limited syntax/semantic manipulation capability. Results are reported in Table~\ref{tab:IBP-vs-vanilla}.

\subsection{Accuracy is a red herring: the BERT case}
We analyse the relationship between semantic robustness and accuracy of a Masked Language Model (MLM): while it is known that MLMs have an improved accuracy on out-of-distribution (OOD) 
data \cite{hendrycks2020pretrained}, there is no clear agreement on the nature of the semantic phenomena, i.e., whether they are \textit{linguistic outliers} or OODs. Although in deep learning a trade off has been observed between the classical notions of robustness and accuracy~\cite{tsipras2018robustness}, 
semantic robustness does not seem to exacerbate this phenomenon. We fine-tuned the BERT language model~\cite{devlinbert:2018} on the SST-2 dataset and tested its robustness on the linguistic phenomena we introduced in the previous section. 
\newline
Despite an accuracy of $0.90$, which is in line with the accuracy of the (simpler) architectures we tested previously, BERT's semantic robustness is 
considerably higher than the "shallow" counterparts (BERT has 16 hidden layers, the models in our benchmark 3). 
BERT has an accuracy of $0.7928$ on \textit{shallow negation}, $0.6974$ on \textit{mixed sentiment}, and $0.8445$ on \textit{sarcasm}\footnote{Due to the high computational cost of fine-tuning BERT, we could not carry out an extensive evaluation correlated by an std interval, as done for the simpler networks.}.
The linguistic phenomenon where BERT performs worst is \textit{mixed sentiment}, as: (i) a few recent works point out the limitations of MLM models such as BERT when learning complex syntactic/semantic constructs~\cite{sinha2021masked}; (ii) we have shown in our previous evaluation that self-attention (along with any other model) is especially brittle to that linguistic construct, despite the layer's name suggesting the opposite. In general, we interpret this linguistic performance as a result of the huge amount of unsupervised training (i.e., the masked language prediction) to which BERT is subjected before being fine-tuned on our supervised task: in this sense, the phase of pre-training, which shapes the dynamics of BERT's contextual embeddings, enables it to considerably outperform shallow models on the linguistic phenomena. 
\newline
We finally validate the results of~\cite{barnes-etal-2019-sentiment}, proving that on their challenging dataset, which contains texts from other non-movie-review datasets (so certainly out of distribution samples), BERT has an accuracy of $0.8552$, $0.6024$ and $0.7111$ on respectively \textit{shallow negation}, \textit{mixed sentiment} and \textit{sarcasm}. This therefore justifies that the task that we set up with our synthetic augmentation through templates is a solid alternative benchmark for semantic robustness. We summarize the results in Table~\ref{tab:bert}.

\begin{figure}[t]
    \centering
    \includegraphics[height=0.5\linewidth,width=0.7\linewidth]{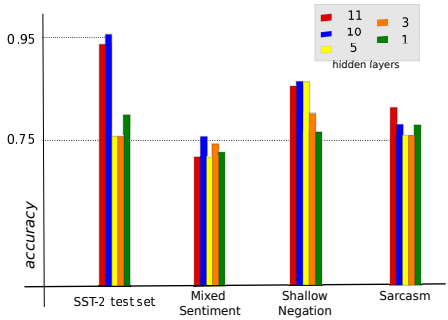}
    \caption{Ablation study of BERT on \cite{barnes-etal-2019-sentiment}, measuring accuracy for 5 different network depths. While depth plays a fundamental role in achieving accuracy on a test set (\textit{SST-2}), and certainly plays a role (albeit minor) on \textit{shallow negation}, it seems not to be correlated to the model performance on \textit{mixed sentiment} and \textit{sarcasm}.}
    \label{fig:ablation-bert}
\end{figure}

\paragraph{Ablation Study of BERT.}
We performed an ablation study of BERT to assess the role of the stacked embeddings to semantic robustness. We hence trained different semantic classifiers on top of a decreasing number of BERT embedding layers. We then measured the semantic robustness on \textit{shallow negation}, \textit{mixed sentiment} and \textit{sarcasm} on samples from \cite{barnes-etal-2019-sentiment}: we found that, despite the accuracy on the task (SST-2) being strongly correlated with the depth of the BERT embedding, semantic robustness is not, as depicted in Figure \ref{fig:ablation-bert}. While the best performing layer is the penultimate? (an interesting phenomenon that is already known in the literature \cite{rogers2020primer}), we could not find a layer that performed the best on all the tasks, a result that leads us to conclude that stacked attention embeddings are fundamental but their internal representation w.r.t. linguistic phenomena (i.e., the `semantics of BERT') is still poorly understood.
To complement the analysis, we tried to disentangle the role of pre-training from that of the embedding depth and attention (which are considered in the design of each BERT hidden layer) by training a very deep LSTM, with 100 input words and an embedding size of $100$, which we then tested on the same semantic phenomena as in the previous evaluation. Interestingly, despite an accuracy of $0.9$ on the SST-2 test set, the accuracy on \textit{shallow negation} is $0.5789$, $0.6684$ on \textit{mixed sentiment} and $0.7$ on \textit{sarcasm}. Although we cannot conclude anything definite, we suspect that the role played by massive pre-training (next word/sentence prediction) is much more important than that of depth and attention, which is in agreement 
observations emerging from other recent studies \cite{liu2021pay}.

\begin{figure}[t]
    \centering
    \includegraphics[height=0.7\linewidth,width=0.7\linewidth]{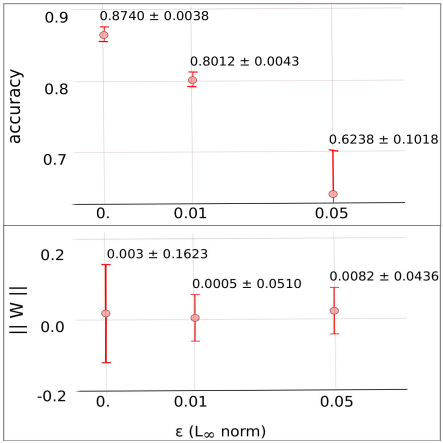}
    \caption{On the top plot, we show average accuracy of $30$ trained FC models on the SST-2 dataset, compared for different values of $\epsilon$-robustness (w.r.t. the $L_{\infty}$-norm). For $\epsilon$ equal to $0.$, a model is not robustly trained, and otherwise it is through IBP~\cite{gowal2018effectiveness}. There is clear a trade-off between robustness and accuracy. On the bottom plot, the average norm of the models' parameters indicates that robust models tend to have lower variance, and hence arguably lower complexity.}
    \label{fig:norm-and-accuracy}
\end{figure}

\begin{table*}[t]
\centering
\begin{adjustbox}{width=2\columnwidth,center}
\begin{tabular}{| *{6}{c|} }
    \hline
    &  \textbf{Train} &\multicolumn{2}{c|}{\textbf{FCs}}
            & \multicolumn{2}{c|}{\textbf{CNNs}}                \\
    \hline
\textit{\textbf{\begin{tabular}[c]{@{}l@{}}Shallow Negation\\ \small{(Our benchmark, Barnes et al., 2019)}\end{tabular}}}   &   \textit{\begin{tabular}[c]{@{}l@{}}Vanilla\\ IBP ($\epsilon=0.001$)\\ IBP ($\epsilon=0.01$)\end{tabular}}  &     \begin{tabular}[c]{@{}l@{}}$0.4034 \pm 0.0214$\\ $0.3852 \pm 0.0071$ \\ $0.4249 \pm 0.0260$ \end{tabular}  &     \begin{tabular}[c]{@{}l@{}} $0.6303 \pm 0.0231$\\ $0.6461 \pm 0.0039$ \\ $0.6145 \pm 0.0263$ \end{tabular}  &     \begin{tabular}[c]{@{}l@{}}$0.3753 \pm 0.0091$\\ $0.4954 \pm 0.0273$* \\ $0.4715 \pm 0.0134$*\end{tabular}  &     \begin{tabular}[c]{@{}l@{}}$0.4553 \pm 0.0719$\\ $0.5079 \pm 0.0822$ \\ $0.4320 \pm 0.0501$\end{tabular}   \\
    \hline
\textit{\textbf{\begin{tabular}[c]{@{}l@{}}Mixed Sentiment\\ \small{(Our benchmark, Barnes et al., 2019)}\end{tabular}}}   &   \textit{\begin{tabular}[c]{@{}l@{}}Vanilla\\ IBP ($\epsilon=0.001$)\\ IBP ($\epsilon=0.01$)\end{tabular}}  &     \begin{tabular}[c]{@{}l@{}}$0.4707 \pm 0.0360$*\\ $0.2918 \pm 0.0121$ \\ $0.2824 \pm 0.0169$ \end{tabular}  &     \begin{tabular}[c]{@{}l@{}} $0.6976 \pm 0.0126$\\ $0.7205 \pm 0.0048$ \\ $0.7072 \pm 0.0133$ \end{tabular}  &     \begin{tabular}[c]{@{}l@{}}$0.4764 \pm 0.0327$\\ $0.5402 \pm 0.0961$ \\$0.4485 \pm 0.0844$ \end{tabular}   &     \begin{tabular}[c]{@{}l@{}} $0.5506 \pm 0.1476$\\ $0.4590 \pm 0.1205$ \\ $0.5506 \pm 0.1476$ \end{tabular}    \\
    \hline
\textit{\textbf{\begin{tabular}[c]{@{}l@{}}Sarcasm\\ \small{(Our benchmark, Barnes et al., 2019)}\end{tabular}}}   &  \textit{\begin{tabular}[c]{@{}l@{}}Vanilla\\ IBP ($\epsilon=0.001$)\\ IBP ($\epsilon=0.01$)\end{tabular}}  &     \begin{tabular}[c]{@{}l@{}}$0.5136 \pm 0.0504$\\ $0.4333 \pm 0.0092$ \\ $0.4406 \pm 0.0943$\end{tabular}  &     \begin{tabular}[c]{@{}l@{}} $0.7133 \pm 0.0156$*\\  $0.5578 \pm 0.0185$ \\ $0.5222 \pm 0.0995$\end{tabular}  &     \begin{tabular}[c]{@{}l@{}}$0.4799 \pm 0.0393$*\\ $0.6352 \pm 0.3962$ \\ $0.1650 \pm 0.1866$ \end{tabular}  &     \begin{tabular}[c]{@{}l@{}} $0.3067 \pm 0.2883$\\ $0.5778 \pm 0.3564$ \\ $0.1593 \pm 0.1030$ \end{tabular}   \\
    \hline
\end{tabular}
\end{adjustbox}
\caption{Comparison of 20 
\textit{IBP-trained robust} models~\cite{gowal2018effectiveness} and their \textit{vanilla} counterparts on samples generated through templates on our benchmark (left subcolumn) and samples exhibiting the same linguistic phenomenon from~\cite{barnes-etal-2019-sentiment} (right subcolumn): both populations of networks have been trained on the SST-2 dataset. IBP, which we use to train robust models for two different values of $\epsilon$ ($0.001$ and $0.01$), cannot ensure robustness to simple semantic rules and in a few cases worsens the performance of the classifier. Symbol \textbf{*}, when present, means that the improved performance (from \textit{vanilla} to \textit{IBP} or vice-versa) is statistically significant. We consider the two architectures (FCs and CNNs) supported by ~\cite{gowal2018effectiveness}.
}
\label{tab:IBP-vs-vanilla}
\end{table*}

\subsection{Robustness Induced Biases}
In this section we examine the relationship between common inductive biases that have inspired the design of machine learning algorithms for the past decades~\cite{mitchell1980need}, and recently also neural networks~\cite{kharitonov2020they}, connecting them to the notions of robustness we dissected in the previous section. In particular, we compare local continuous to local semantic robustness.

\paragraph{Minimum Cross-validation Error.} There is empirical evidence in the literature~\cite{huang2019achieving,jia2019certified} that continuous robustness does not naturally induce better performance on trained models. Indeed, most of the models that are trained to be robust are less accurate than the brittle counterparts. This side-effect is caused by the margin that is propagated through the network to the output to induce invariance to nearest neighbours of a given input. ``Shielding'' the model with a thick margin of possibly unrelated terms leads to an inconsistent treatment of different sentences (as noted in Observation 2, human language abounds in edge cases). This is testified by further experiments shown in Figure~\ref{fig:norm-and-accuracy} (top). Concerning semantic robustness, generalization on cogent linguistic rules does not necessarily benefit a model's performance, as demonstrated by experiments we conducted on 30 networks trained to be semantically robust against \textit{shallow negation} vs. their vanilla counterpart: both populations have been trained on the SST-2 dataset~\cite{socher2013recursive}. Robustness is enabled through simple data augmentation on the dataset provided by Barnes et al.~\cite{barnes-etal-2019-sentiment}, whereas the test is performed on unseen sentences that exhibit the same linguistic phenomenon. 
While the vanilla networks have an average accuracy of $0.9036 \pm 0.0019$ on the test set and $0.4916 \pm 0.0074$ on the \textit{shallow negation} test set, those that have been robustly trained have an accuracy of $0.8838 \pm 0.0049$ and $0.5491 \pm 0.0124$, respectively. 

\paragraph{Minimum Description Length.} Local continuous robustness is known to be a strong regularizer~\cite{gowal2018effectiveness}. In fact, classical methods used to induce local robustness for NLP (such as IBP), 
which propagate through all the embedding dimensions and thus amplify the noise,  nonetheless play an important role as they smooth out the network's hidden activations. We report the results of experiments that we conducted that support this hypothesis in~Figure~\ref{fig:norm-and-accuracy} (bottom). As regards semantic robustness, we cannot conclude anything definitive but the evidence suggests that semantically robust models are not necessarily smoother than the vanilla counterparts. We compared the weights' norm of 30 networks trained to be robust against \textit{shallow negation} vs. their vanilla counterparts (see previous paragraph for details). While the difference between the performance of the two networks on unseen texts that contain that linguistic phenomenon is substantial, there is very little difference in the norm of the two populations, which are respectively $0.0017 \pm 0.0019$ (vanilla) and $0.0064 \pm 0.0032$ (robust).

\paragraph{Nearest Neighbours.} 
Local robustness induces a strong bias towards nearest neighbours, by definition. This assumption is critical as robust training underestimates the effect of making a model robust, treating all the dimensions in the embedding as equally important. We hypothesize this causes the 
deterioration of 
the performance of robust models in NLP. The induced invariance along any dimension reduces the effectiveness of the embedding representation on cogent syntactic/semantic tasks such as word-sense-disambiguation, polysemy, etc. 
\textit{Semantic robustness} takes a different approach and is expected not to be robust to nearest neighbours in the embedding space, but rather to perturbations that are generated by the linguistic rules for which they have been robustly trained. For an increasing number of embedding dimensions, semantic robustness does not suffer in principle from the trade-off between the performance on linguistic tasks \cite{chen2013expressive} and robustness guarantees \cite{la2020assessing}.

\section{Conclusions}
In this paper we formalise the concept of \textit{semantic robustness}, which generalizes the notion of NLP robustness by explicitly considering the measurement of robustness on cogent linguistic phenomena. We propose a template-based generative test bed to evaluate semantic robustness. 
We conduct an empirical analysis that demonstrates that, despite being harder to implement, \textit{semantic robustness} provides stronger guarantees for complex linguistic phenomena where models robust in the classical sense fail. In future, we aim to automate, when possible, the generation of semantic test beds, aided by powerful Masked Language Models such as GPT. We further plan to introduce a validation step for the newly generated texts by involving humans to assess the quality of the semantic perturbations (and consequently of the semantic rules). Finally, we aim to extend our analysis to high-order NLP tasks and
study the relationship of linguistic phenomena with out-of-distribution and outlier samples.  

\section{Acknowledgements}
This project has received funding from the European Research Council (ERC)
under the European Union’s Horizon 2020 research and innovation programme
(FUN2MODEL, grant agreement No.~834115).

\bibliographystyle{abbrv}
\bibliography{bib}

\end{document}